\documentclass{article}
\usepackage[preprint]{corl_2026}

\usepackage{times}

\usepackage{multicol}

\usepackage{graphicx}
\usepackage{amsmath}
\usepackage{amssymb}
\usepackage{subcaption}
\usepackage{stfloats}
\usepackage{wrapfig}
\usepackage{booktabs}
\usepackage{threeparttable}
\usepackage{multirow}
\usepackage{placeins}

\makeatletter
\@ifpackageloaded{natbib}{}{\usepackage[numbers]{natbib}}
\@ifpackageloaded{hyperref}{}{\usepackage[bookmarks=true]{hyperref}}
\makeatother

\usepackage{fancyhdr}

\fancypagestyle{acceptedfirstpage}{
  \fancyhf{}
  \fancyhead[L]{\small\itshape Accepted to CoRL 2026}
  
}

\begin{document}

\title{Mind the Gap: Rethinking I/O Design for Contact-Rich Visuomotor Policy Learning}

\author{
{\bfseries Cuijie Xu$^1$, Shurui Zheng$^{1,2}$, Zihao Su$^3$, Zhongchen Jian$^1$, Yuanfan Xu$^4$,} \\
{\bfseries Tinghao Yi$^5$, Xudong Zhang$^1$, Jian Wang$^1$, Yu Wang$^1$, Jinchen Yu$^{1*}$} \\[0.8ex]
\normalfont\small
$^1$Department of Electronic Engineering, Tsinghua University \\
$^2$Qiuzhen College, Tsinghua University \\
$^3$School of Mechanical and Aerospace Engineering, Jilin University \\
$^4$Striding AI \\
$^5$EFORT Intelligent Robot Co., Ltd. \\
$^*$Email: \texttt{yu-jc@mail.tsinghua.edu.cn}
}

\maketitle
\thispagestyle{acceptedfirstpage}

\begin{abstract}
Contact-rich teleoperation logs expose a policy I/O design choice: demonstrations may contain the robot execution ($E$), leader command ($C$), or both. These signals are not interchangeable: E2E may discard contact-generating command offsets, whereas E2C preserves these offsets but omits the robot's execution response. We propose Dual-State Conditioning (EC2C), which conditions on both $E$ and $C$ while predicting future $C$, exposing command-execution mismatch as a cue for contact, latency, payload, and operator compensation; in quasi-static contact, this cue is often force-correlated. On a low-cost setup without force, tactile, or motor-current policy input, EC2C outperforms E2E and a strong E2C baseline across several real-world contact-rich, force-sensitive, and dynamic tasks. These results support EC2C as a practical default I/O setting for contact-rich imitation learning. We further formulate latency-adaptive inpainting as a temporal extension of this I/O choice for action-chunking policies, and discuss when long histories help dynamic inference or introduce causal confounding. 
\end{abstract}

\section{Introduction}

Visuomotor imitation learning is often framed as mapping observations to actions. In teleoperated manipulation, however, choosing the action target and proprioceptive input is itself part of the system design. A demonstration may record the robot execution, the operator or leader command, or both. These signals are not interchangeable when the low-level controller and hardware are imperfect.

Low-cost robots are rarely ideal actuators. Communication latency, finite controller gains, backlash, friction, gravity compensation error, and contact all create a gap between the command sent to the robot and the motion that is physically executed. Human teleoperators compensate for this gap, but imitation learning pipelines often collapse the resulting data into a single state-action stream.

\begin{figure}[htbp]       
    \centering
    \includegraphics[width=0.60\linewidth,trim=0 40bp 0bp 0bp,clip]{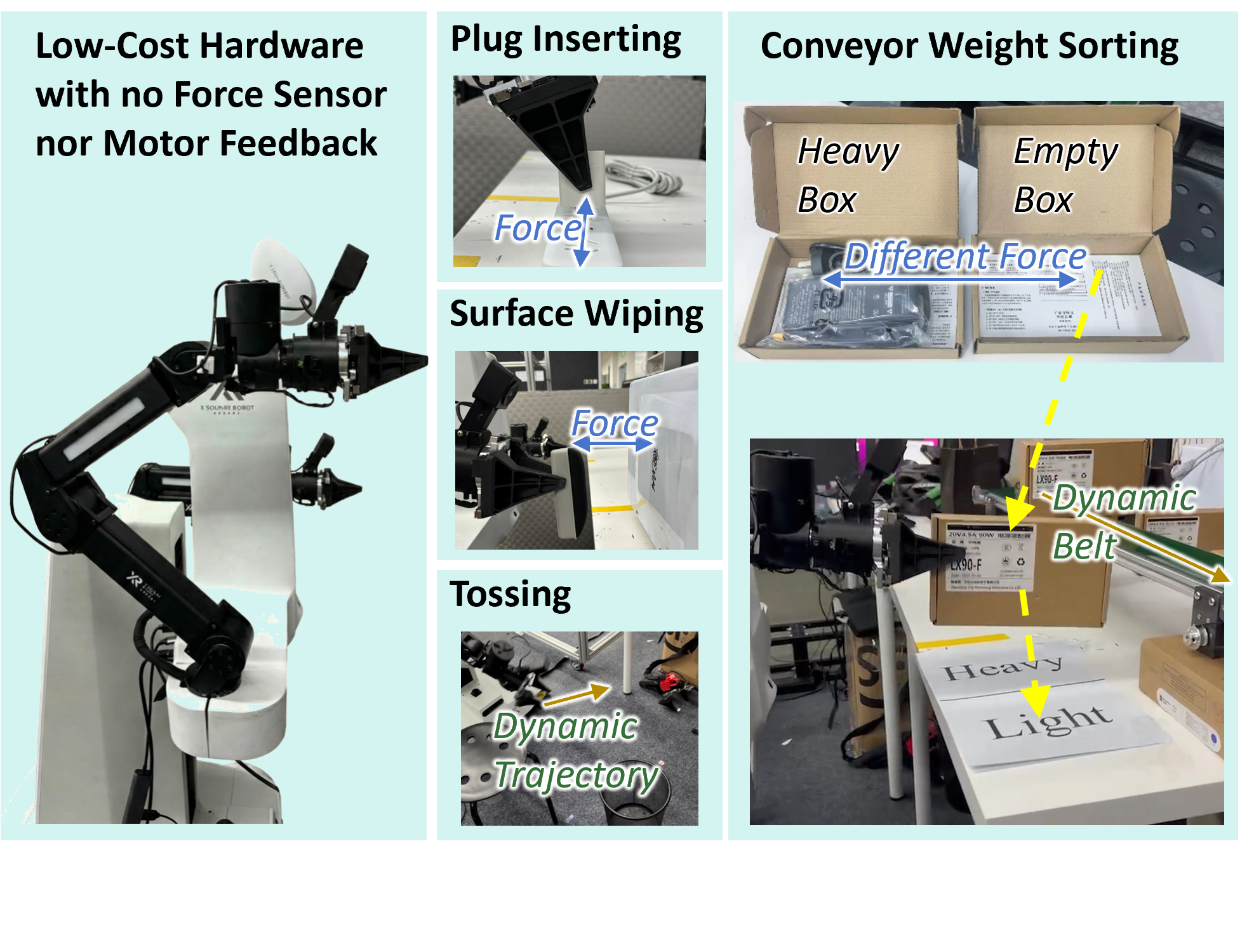}
    \caption{\textbf{Mismatch-Aware Imitation Learning on Sensorless Hardware.} Our method enables low-cost, position-controlled robots (left) to perform force-sensitive and dynamic tasks without explicit force, tactile, or motor current feedback. By conditioning on command-execution mismatch, the policy succeeds on high-precision plug insertion, force-modulated surface wiping, dynamic tossing, and proprioceptive weight sorting of visually identical objects.}
    \label{fig:fig0}
    \vspace{-20pt}
\end{figure}

In the sensorless, force-sensitive manipulation setting shown in Fig.~\ref{fig:fig0}, we ask: \textbf{how should contact-rich visuomotor policies choose their input and output signals from demonstrations collected on imperfect low-cost hardware?}

Demonstration pipelines expose different I/O choices. UMI-style \cite{chi2024universalmanipulationinterfaceinthewild} and ego-centric collection \cite{shaw2023givingrobotshand,kareer2024egomimicscalingimitationlearning,zheng2026egoscalescalingdexterousmanipulation} can scale demonstrations from execution-like trajectories, wrist motion, or retargeted hand actions, yielding execution cloning (E2E) when no separate command channel exists. Leader-follower and force-feedback teleoperation additionally record the leader command ($C$), which encodes operator intent, and commonly support command cloning (E2C): predicting leader-side commands or command-like action chunks from follower-side observations \cite{zhao2023learning,fu2024mobilealohalearningbimanual,tsumugi2020imitation,fukuda2024biact,namekawa2024ilbit,kobayashi2025bilat}.

This I/O view exposes complementary failure modes. E2E is appropriate when only execution labels are available, but on imperfect hardware it can remove the command offsets that operators use to sustain contact or overcome steady-state error. E2C preserves these compensatory commands, but from execution history alone it does not expose whether the same command is tracking freely, being resisted by contact, or changing under payload and controller error. When paired command and execution traces are available, this leaves an unused relation: how the robot responded to the command under the current controller, hardware, and contact condition.

We introduce Dual-State Conditioning (EC2C): condition on both execution $E$ and command $C$ while predicting future $C$. EC2C preserves the command target while making command-execution mismatch available as an execution-grounded cue. Although this cue is not a replacement for force or tactile sensing, it provides a sensorless, force-correlated signal in quasi-static contact using only the command and execution traces already present in the log. More generally, the mismatch reflects the response of the controller, hardware, environment, and operator. For large action-chunking policies, this I/O choice also has a temporal extension. During inference latency, the robot continues executing a previously generated command buffer, so part of the future command sequence is already committed and available for conditioning; the corresponding future execution, however, is not yet observable. We use latency-adaptive inpainting to condition generation on this committed command prefix, reducing the train-inference gap and maintaining smooth asynchronous execution.

Our contributions are:

\textbf{I/O-Space Deconstruction:} We formulate contact-rich imitation learning as an I/O design problem and compare E2E, E2C, and EC2C under the same low-cost hardware and policy backbones. This view provides a common language for relating execution-centric data collection, command cloning, and controller compensation.

\textbf{Dual-State Conditioning as a Default:} We propose EC2C as a default I/O configuration: it keeps the command target used in E2C while adding the recorded command history to the policy input, making command-execution mismatch available at no extra sensing cost. We characterize when this cue is force-correlated, when it reflects broader controller dynamics, payload, friction, latency, or operator effects, and why long histories can introduce causal confounding in quasi-static tasks.

\textbf{Latency-Adaptive Deployment:} We formulate inpainting as a temporal extension of EC2C: committed future commands are available during inference latency, while future execution is not yet observable. Conditioning on this command prefix reduces the train-inference gap and supports smooth asynchronous action-chunk execution.

\textbf{Broad Empirical Validation:} We validate the universality of our framework by instantiating it across diverse backbone architectures, including Diffusion Policy \cite{chi2023diffusion} and $\pi_0$ \cite{black2024pi0}. Through extensive experiments ranging from contact-rich to dynamic manipulation, we demonstrate that our approach consistently achieves robust control on sensorless hardware.

\section{Related Work}
\subsection{Visuomotor Imitation Learning and I/O Spaces}
Visuomotor imitation learning has evolved from one-step behavior cloning toward action-chunking policies such as ACT \cite{zhao2023learning}, generative models such as Diffusion Policy \cite{chi2023diffusion}, and generalist robot policies such as RT-2, Octo, RT-X, and $\pi_0$ \cite{brohan2023rt2,octomodelteam2024octoopensourcegeneralistrobot,open2023open,black2024pi0}. These works study scalable architectures, data, action sequence modeling and representations. Our focus is complementary: which logged physical signals should define the policy's default I/O setting, including its inputs and action target.

Demonstration interfaces determine which physical signals are available. UMI~\cite{chi2024universalmanipulationinterfaceinthewild} and ego-centric demonstration methods~\cite{kareer2024egomimicscalingimitationlearning,zheng2026egoscalescalingdexterousmanipulation} provide execution-like targets, corresponding to E2E in our taxonomy. Leader-follower systems such as ALOHA~\cite{zhao2023learning,fu2024mobilealohalearningbimanual} and bilateral imitation systems ~\cite{tsumugi2020imitation,fukuda2024biact,namekawa2024ilbit,kobayashi2025bilat} record leader commands as action targets, making command cloning (E2C) a strong baseline. SM2SM autoregressive models~\cite{sasagawa2021motion} condition on and predict both master and slave states, a paired-state design close to EC2C. Our focus differs in problem setting and deployment: we study contact-rich visuomotor action-chunking on imperfect low-cost hardware. EC2C keeps command prediction as the action target while exposing command-execution mismatch as policy context, without force, tactile, or motor-current input.

\subsection{Force-Aware and Sensorless Contact-Rich Learning}
Contact-rich manipulation often benefits from physical feedback. One line of work studies how to use these signals in learned policies, including visual-tactile diffusion policies \cite{xue2025reactivediffusionpolicyslowfast,chen2025implicitrdpendtoendvisualforcediffusion}, force/torque-aware VLA modules \cite{yu2025forcevlaenhancingvlamodels,li2026favlaforceadaptivefastslow,zhang2025tavlaelucidatingdesignspace}, and ACT-style policies with external-torque input \cite{sujit2025improvinglowcostteleoperation}. These methods use measured force signals as policy inputs or auxiliary targets. Another line avoids dedicated force/torque or tactile sensors, but still explicitly estimates contact force through dynamics models. Classic observer-based methods estimate external torques from robot dynamics and proprioceptive measurements \cite{garofalo2019slidingmodemomentumobservers,sariyildiz2023stabilityanalysisreactiontorque}; recent sensorless bilateral teleoperation applies this idea to low-cost hardware with identified dynamics and a disturbance observer \cite{yamane2025designexperimentalvalidation}. EC2C adds no physical sensor and does not explicitly estimate force. Instead, it uses the position-level command-execution mismatch already present in paired teleoperation logs. In quasi-static contact, this mismatch can be force-correlated; outside that regime, it remains a broader cue for tracking error, payload, latency, friction, and operator compensation.

Recent teleoperation work also recognizes that command-execution mismatch can provide useful contact feedback during data collection. ACE-F \cite{yan2025acefcrossembodiment} renders virtual force from commanded and actual end-effector positions; we instead use the same type of signal as policy input.

\subsection{Action-Chunking Smoothing and Temporal I/O}
Action-chunking policies must smooth transitions between chunks. ACT \cite{zhao2023learning} uses temporal ensembling, while RTC \cite{black2025realtimeexecutionactionchunking} treats asynchronous execution as inference-time inpainting with a fixed latency prefix. Training-time RTC \cite{black2025trainingtimeactionconditioning} learns a suffix model $p(C_{t+d:t+H}\mid o_t, C_{t:t+d})$ by masking prefix losses and hard-clamping during sampling. Our latency-adaptive inpainting is a simpler conditioning-only variant: we keep the original training and sampling pipeline while conditioning the chunk model $p(C_{t:t+H}\mid o_t, C_{t:t+d})$ on committed commands from the previous prediction, and discard the generated prefix at deployment. It is EC2C's natural temporal extension: the command prefix is already known, but the corresponding future execution is not.

\section{Methodology}

\subsection{I/O Spaces under Imperfect Controllers}
\label{sec:problem_formulation}

We formalize visuomotor control in task space. Let $I_t$ denote the visual observation, $\mathbf{x}^e_t \in \mathcal{X}$ denote the \textit{follower execution} state, and $\mathbf{x}^c_t \in \mathcal{X}$ denote the \textit{leader command} state. The state includes the end-effector pose and gripper state when applicable. We use $\mathbf{X}^{e}_{a:b}$ and $\mathbf{X}^{c}_{a:b}$ to denote execution and command sequences, with $L$ for observation history length and $H$ for prediction horizon.

\textbf{Dynamics of the Black-Box Controller.}
In low-cost or compliant setups, the command target and executed motion can differ systematically. We model the low-level control stack, including IK, PID, latency, and environmental contact, as a black-box map $\mathcal{B}$ under physical condition $\theta_t$:
\begin{equation}
    \mathbf{x}_{t+1}^e = \mathcal{B}(\mathbf{x}_{t}^e, \mathbf{x}_{t}^c; \theta_t)
\end{equation}
This generalized model captures various system discrepancies driven by $\theta_t$, including finite controller stiffness, latency, friction, gravity and payload, all of which can make $\mathbf{x}^e$ deviate from $\mathbf{x}^c$.

\begin{figure}[htbp]
    \centering
    \begin{minipage}[t]{0.49\textwidth}
        \vspace{0pt}
        \centering
        \includegraphics[width=0.96\linewidth]{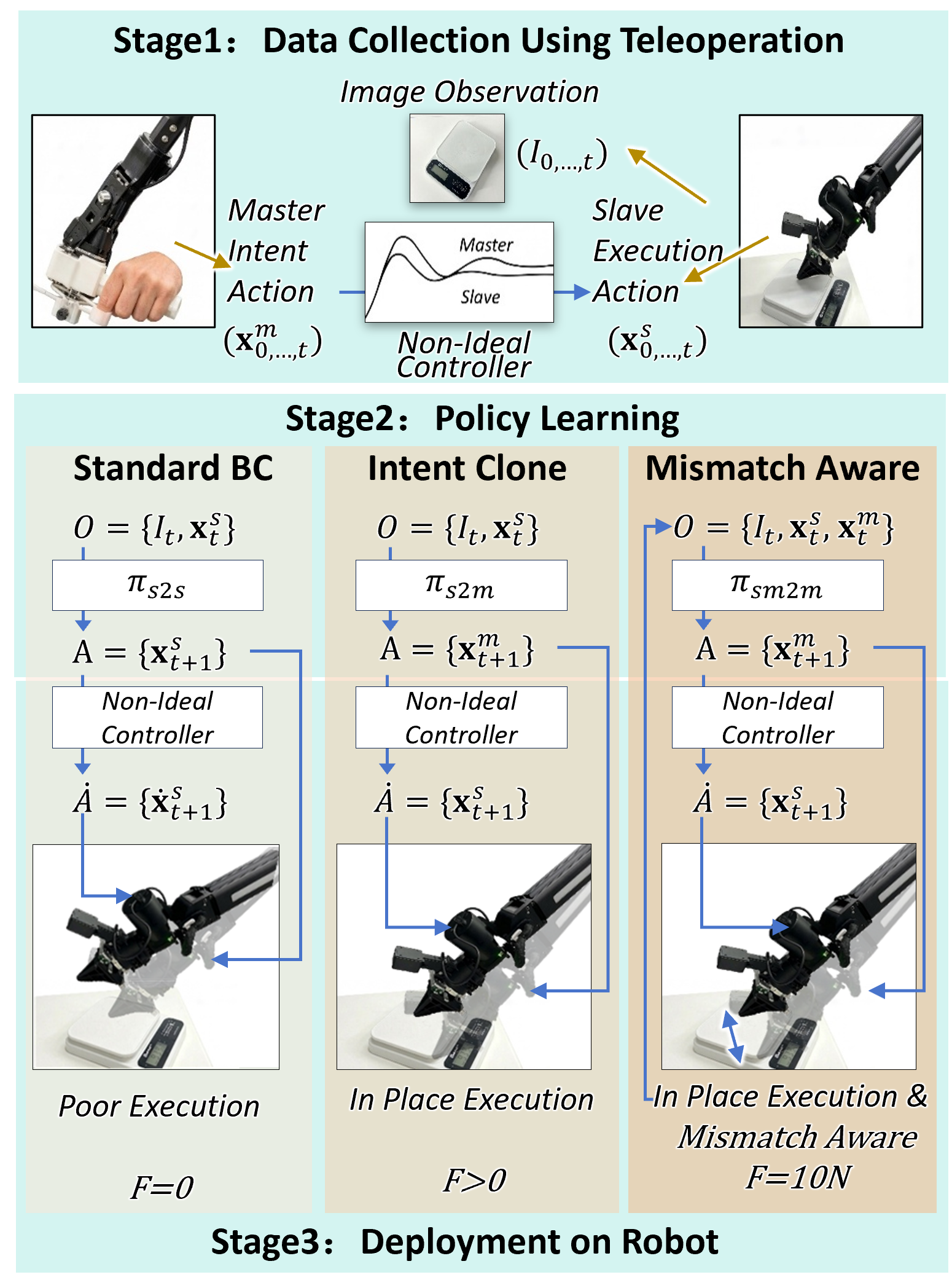}
        \caption{\textbf{I/O choices from paired teleoperation logs.} E2E clones follower execution and can lose contact-producing command offsets. E2C predicts leader commands but lacks the execution response. EC2C conditions on both histories while predicting commands, exposing command-execution mismatch to the policy.}
        \label{fig:fig1}
    \end{minipage}
    \hfill
    \begin{minipage}[t]{0.49\textwidth}
        \vspace{0pt}
        \centering
        \includegraphics[width=0.96\linewidth]{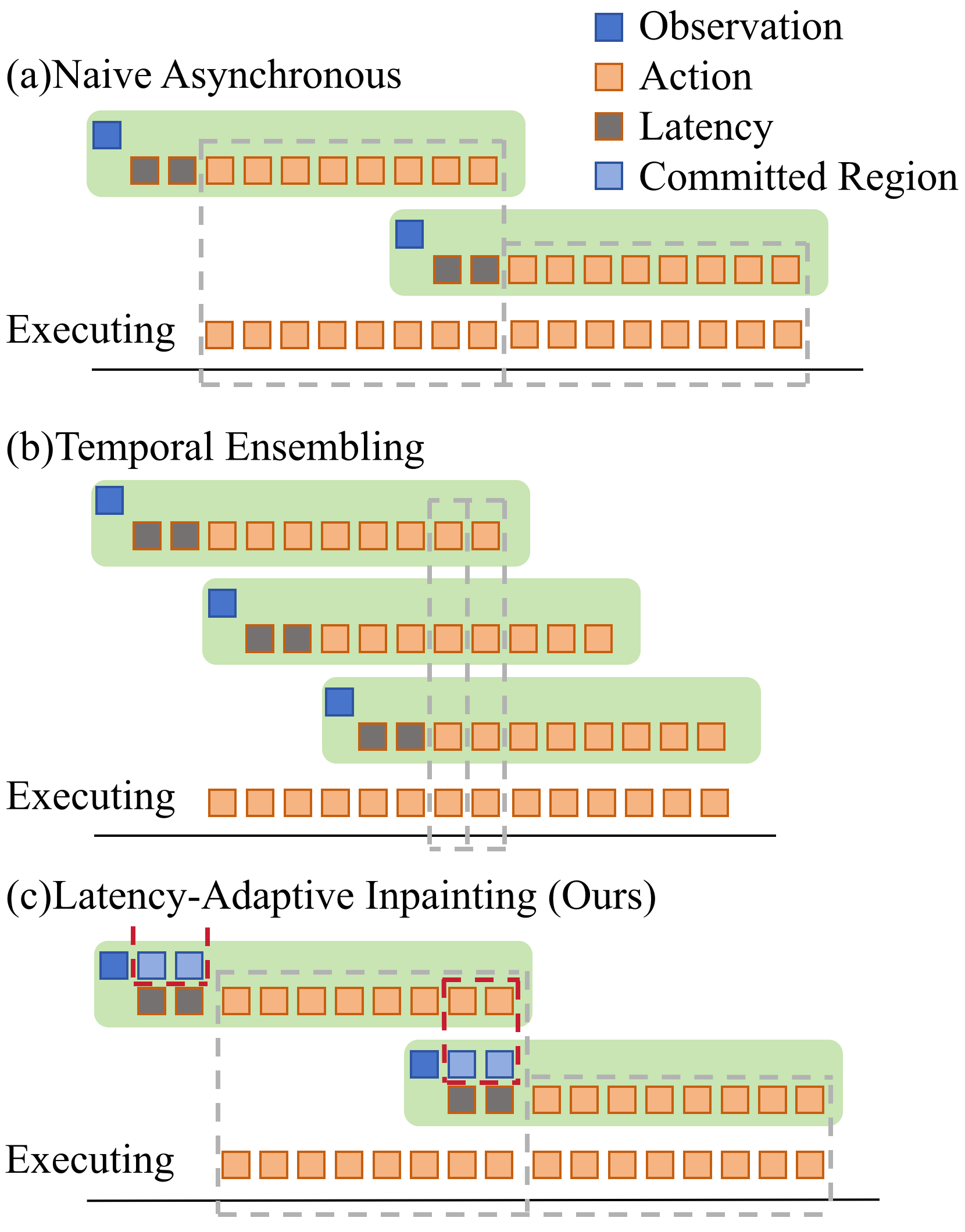}
        \caption{\textbf{Comparison of asynchronous control pipelines.} Naïve streaming can create jumps; Temporal Ensembling smooths by averaging stale predictions but introduces reaction lag; latency-adaptive inpainting conditions on the committed command prefix for continuity while maintaining immediate reactivity to the latest observation.}
        \label{fig:inpainting}
    \end{minipage}
\end{figure}

\textbf{Human Compensation in Teleoperation.}
Teleoperation data is a paired trace of the operator command and the execution produced by the hardware and environment. To achieve desired object motion $\mathbf{x}^{des}$, the operator often commands a leader state that differs from the desired follower state:
\begin{equation}
    \mathbf{x}_{t}^c = \mathbf{x}_{t+1}^{des} \oplus \Delta_t(\mathbf{x}_{t}^{e}, \theta_t),
    \qquad
    \boldsymbol{\epsilon}_t = \mathbf{x}^c_t \ominus \mathbf{x}^e_t .
\end{equation}
Here $\Delta_t$ summarizes operator compensation, and $\boldsymbol{\epsilon}_t$ is the \emph{command-execution mismatch}. When exposed by the I/O choice, this mismatch becomes an input feature and a composite trace of controller dynamics, contact, latency, friction, gravity, payload, and operator compensation.

\textbf{When Mismatch is Force-Correlated.}
In the special case of local quasi-static motion, rigid or locally linear contact, stable position control, and operation away from saturation, the external interaction wrench becomes correlated with mismatch:
\begin{equation}
    \mathbf{F}_{ext} \approx \mathcal{F}(\boldsymbol{\epsilon}_t, \mathbf{x}^e_t) \approx K_x(\mathbf{x}^e_t)\boldsymbol{\epsilon}_t
\end{equation}

 We characterize this approximation in Sec.~\ref{sec:local_force_correlation}. It supports force-related behavior such as pressing or wiping. Thus, in quasi-static contact-rich settings, EC2C exposes a useful signal already present in paired logs under the same low-cost controller used at deployment.

\textbf{Deconstructing Learning Objectives.}
The paired log yields three I/O choices, illustrated in Fig.~\ref{fig:fig1}:
\begin{equation}
\begin{aligned}
    \pi_{\theta}^{\mathrm{E2E}}(I_t, \mathbf{X}^{e}_{t-L:t}) &\rightarrow \mathbf{X}^{e}_{t+1:t+H},\\
    \pi_{\theta}^{\mathrm{E2C}}(I_t, \mathbf{X}^{e}_{t-L:t}) &\rightarrow \mathbf{X}^{c}_{t+1:t+H},\\
    \pi_{\theta}^{\mathrm{EC2C}}(I_t, \mathbf{X}^{e}_{t-L:t}, \mathbf{X}^{c}_{t-L:t}) &\rightarrow \mathbf{X}^{c}_{t+1:t+H}.
\end{aligned}
\end{equation}
\textbf{E2E} fits execution-only logs or near-ideal controllers, but on imperfect hardware it can remove command offsets used for contact pressure or steady-state error compensation. \textbf{E2C} preserves the operator's command target, including virtual-equilibrium offsets, but does not observe whether the command produced the intended execution. \textbf{EC2C} keeps E2C's deployed action target and adds command history, making mismatch observable. In quasi-static tasks, using the current or very recent mismatch avoids confounding task phase with contact state. In dynamic tasks, a short history can reveal temporal effects such as delayed tracking.

\subsection{Latency-Adaptive Control via Sequence Inpainting}

As shown in Fig.~\ref{fig:inpainting}, high-capacity action-chunking policies can have inference latency $\delta_{inf}$. During this delay, the robot executes the previous command buffer. Replacing chunks naively can cause discontinuities, while synchronous waiting can freeze the robot during dynamic tasks. We therefore treat the next action chunk as a temporal I/O-conditioning problem.

Let $\mathbf{X}^c_{t+1:t+H}$ be the future command sequence and $d=\lceil \delta_{inf}/\Delta t\rceil$ the committed prefix length. The first $d$ commands are fixed by the previous buffer, while the corresponding future execution is unobserved. We condition on this prefix and still predict a full command chunk:

\begin{equation}
    \pi_{\theta}^{\mathrm{EC2C\text{-}IP}}
    (I_t,\mathbf{X}^{e}_{t-L:t},\mathbf{X}^{c}_{t-L:t},
    \mathbf{X}^{c,\mathrm{buf}}_{t+1:t+d},\mathbf{M})
    \rightarrow \mathbf{X}^{c}_{t+1:t+H}
\end{equation}

where $\mathbf{M}\in\{0,1\}^H$ marks committed steps. During training, we randomize $d\sim\mathcal{U}[0,d_{\max}]$; at inference, $d$ is measured online. We keep the full-chunk objective and sampler unchanged, appending the prefix and mask as conditioning tokens for VLA policies or channels for diffusion policies. At deployment, the generated prefix is discarded and only the newly planned suffix is executed.

This is EC2C's temporal extension: EC2C separates observed execution from command targets in the spatial I/O space; asynchronous deployment creates the same distinction over time. Future commands can be known because they were committed by the previous buffer, but future execution is not observable before it happens. Conditioning on the committed prefix reduces the train-inference gap and encourages smooth continuation without changing the policy objective.

\begin{figure*}[b]
    \centering
    \includegraphics[width=0.9\textwidth]{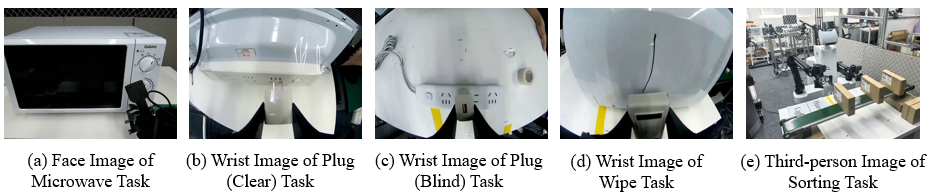}
    \caption{Overview of the manipulation tasks.}
    \label{fig:tasks}
\end{figure*}

\section{Experiment}

We evaluate whether EC2C improves teleoperation imitation learning without force, tactile, or motor-current policy input. The experiments ask whether command-execution mismatch is locally force-correlated (Q1), whether E2C preserves force-producing offsets (Q2), whether EC2C improves over E2C by exposing execution response (Q3), how observation horizon affects quasi-static and dynamic tasks (Q4), whether mismatch improves generalization under system-dynamics shift (Q5), and whether latency-adaptive inpainting maintains smooth asynchronous control (Q6).

\subsection{Setup and Task Design}

We use a low-cost bimanual teleoperation suite with two ARX-R5 follower and ARX-X5 leader 6-DoF arms. The system operates at a control frequency of 60Hz. Policies receive RGB observations from two wrist cameras and one head camera at 20Hz, plus position-level leader/follower states; force, tactile, and motor-current inputs are excluded.

The task suite shown in Fig.~\ref{fig:tasks} is chosen to isolate distinct failure modes. \textbf{Microwave Manipulation} is a long-horizon sequence in which the robot presses a stiff mechanical button, moves an object, and closes the door; the button press stage tests whether the policy preserves contact-generating command offsets rather than cloning the stopped execution pose. \textbf{Plug Insertion (Clear / Blind)} requires inserting a two-prong plug into a tight socket. In the Clear variant, the holes remain visible and the task mainly tests force-producing grasp and insertion commands; in the Blind variant, the plug occludes the holes during contact, so the policy must distinguish faceplate collision from sliding into the socket. \textbf{Surface Wiping} erases ink marks from an unsecured microwave surface, requiring enough normal pressure to clean without pushing the object away. \textbf{Dynamic Tossing} throws an object into a bin beyond the robot's kinematic reach, stressing velocity continuity and release timing under action-chunk latency. \textbf{Conveyor Weight Sorting} grasps moving, visually identical 50g/350g objects and sorts them by mass, combining latency-sensitive tracking with mass inference from command-execution mismatch during lifting.

\textbf{Data collection.}
Long proprioceptive histories can create a shortcut in which the policy predicts the next command from past trends instead of current visual alignment. We collect recovery intervention data inspired by DAgger \cite{ross2011reduction}: a supervisor perturbs the target or path and the operator recovers, weakening history-only correlations. This mitigation is incomplete, so we still tune the observation horizon by task. Our teleoperation system does not render force or tactile feedback to the operator. During data collection, operators therefore use privileged context when needed: side views for contact-heavy tasks such as Blind Plug Insertion and Wiping, and known object mass for Conveyor Weight Sorting. This information is withheld from the policy; the policy receives only its RGB observation and paired command-execution traces.

\subsection{Result}

\subsubsection{\textbf{Local Force Correlation (Q1)}}
\label{sec:local_force_correlation}

\begin{wrapfigure}{r}{0.5\textwidth}
  \vspace{-60pt}
  \centering
  \includegraphics[width=\linewidth]{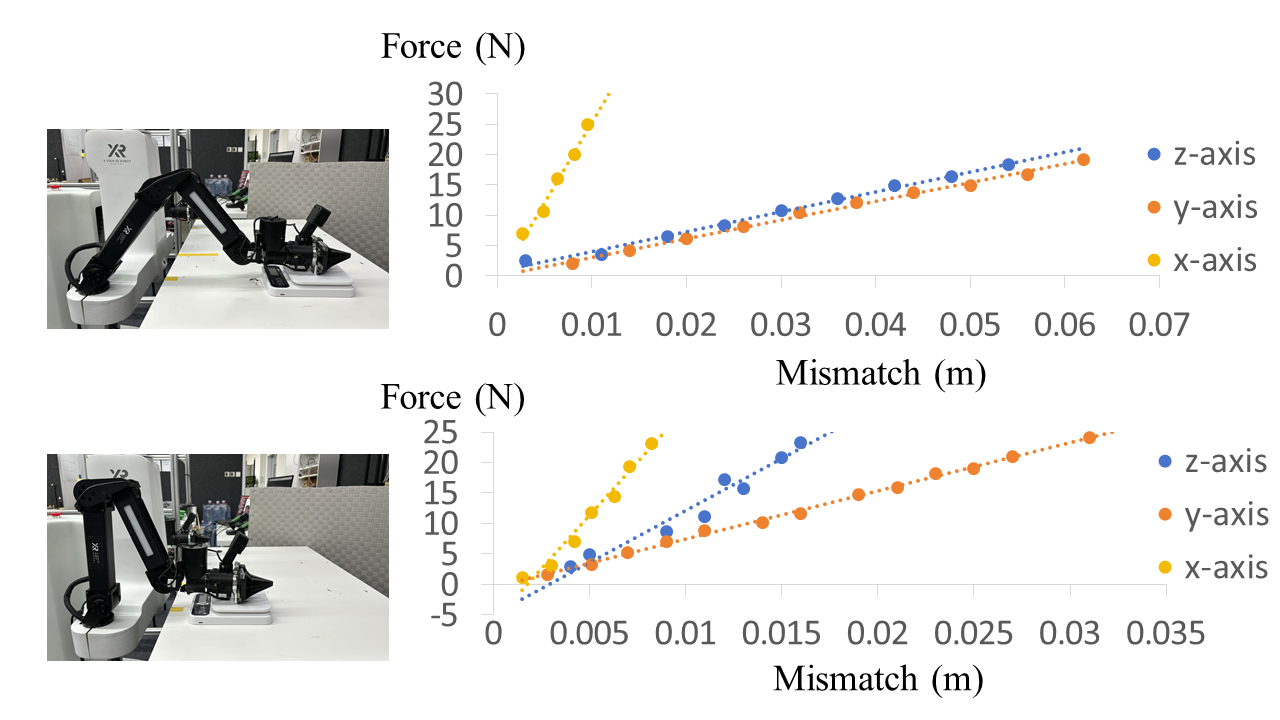}
  \caption{\textbf{Local Force Correlation (Q1).}
  We characterize the relationship between command-execution mismatch ($\boldsymbol{\epsilon}$) and ground-truth interaction force under quasi-static rigid contact across two workspace configurations: (Down) Contracted Pose and (Up) Extended Pose.}
  \label{fig:stiffness_plot}
  \vspace{0pt}
  
  \captionof{table}{\textbf{Stiffness Characterization.} Effective stiffness $k$ (N/m) and linearity ($R^2$) across different axes and robot configurations.}
  \label{tab:stiffness}
  \resizebox{\linewidth}{!}{%
    \begin{tabular}{@{}lccccc@{}}
    \toprule
    & \multicolumn{2}{c}{\textbf{Contracted Pose}} & \multicolumn{2}{c}{\textbf{Extended Pose}} & \\ \cmidrule(lr){2-3} \cmidrule(lr){4-5}
    \textbf{Axis} & $k$ (N/m) & $R^2$ & $k$ (N/m) & $R^2$ & \textbf{Change} \\ \midrule
    X-axis (Longitudinal) & 3501 & 0.981 & 2914 & 0.977 & -17\% \\
    Y-axis (Lateral) & 791 & 0.998 & 314 & 0.995 & -60\% \\
    Z-axis (Vertical) & 1709 & 0.977 & 328 & 0.992 & -81\% \\ \bottomrule
    \end{tabular}%
  }
  \vspace{-20pt}
\end{wrapfigure}

We measure interaction force with a digital scale during controlled indentation. Fig.~\ref{fig:stiffness_plot} and Table~\ref{tab:stiffness} answer Q1 in two parts. First, under quasi-static rigid contact, command-execution mismatch is strongly correlated with interaction force across all tested axes and configurations ($R^2>0.95$), supporting its use as a local force-correlated cue for tasks such as pressing and wiping. Second, the effective stiffness is highly configuration-dependent: the lateral stiffness changes by $60\%$ and the vertical stiffness by $81\%$ between the two tested poses. Thus, a single global stiffness constant is insufficient; the useful signal is the mismatch itself, to be interpreted by the learned policy together with visual and proprioceptive context. In full manipulation tasks, mismatch can also reflect latency, friction, gravity, making it a broader execution-grounded cue for the learned policy.

\begin{table*}[t]
\centering
\caption{\textbf{Manipulation Capability Analysis.} We report success rates across four tasks. For the long-horizon Microwave task, we report the success rate for each stage (Open, Grasp, Place, Close) to diagnose different failure modes. For the Wiping task, we report both Strict Success (100\% erased) and Effective Rate (residue $<50\%$)}
\label{tab:main_results}
\resizebox{0.9\textwidth}{!}{%
\begin{tabular}{ll cccc c c cc}
\toprule
 & & \multicolumn{4}{c}{\textbf{Microwave Sequence}} & \multicolumn{1}{c}{\textbf{Plug (Clear)}} & \multicolumn{1}{c}{\textbf{Plug (Blind)}} & \multicolumn{2}{c}{\textbf{Wiping}} \\
\cmidrule(lr){3-6} \cmidrule(lr){7-7} \cmidrule(lr){8-8} \cmidrule(lr){9-10} 
\textbf{Arch.} & \textbf{Method} & Open & Grasp & Place & Close & Success & Success & Strict & Effective \\
\midrule
\multirow{3}{*}{Diffusion Policy} 
 & E2E & 0/20 & 11/20 & 15/20 & 14/20 & - & - & 0/20 & 0/20 \\
 & E2C & 14/20 & 11/20 & 12/20 & 12/20 & - & - & 4/20 & 9/20 \\
 & \textbf{EC2C (Ours)} & \textbf{20/20} & 12/20 & 17/20 & \textbf{20/20} & - & - & 13/20 & 17/20 \\
\midrule
\multirow{3}{*}{\textbf{$\pi_0$}} 
 & E2E & 0/20 & \textbf{19/20} & 17/20 & 13/20 & 0/20 & 0/20 & 1/20 & 4/20 \\
 & E2C & 18/20 & 18/20 & \textbf{20/20} & 18/20 & \textbf{19/20} & 8/20 & 4/20 & 13/20 \\
 & \textbf{EC2C (Ours)} & \textbf{20/20} & \textbf{19/20} & \textbf{20/20} & \textbf{19/20} & \textbf{19/20} & \textbf{16/20} & \textbf{18/20} & \textbf{20/20} \\
\bottomrule
\end{tabular}%
}
\end{table*}

\subsubsection{\textbf{Command Cloning for Force Generation (Q2)}}

Table~\ref{tab:main_results} shows that E2C recovers much of E2E's loss on force-producing subtasks by preserving operator command offsets. In the microwave task, E2E clones the follower pose that stops at the stiff button surface, while E2C commands beyond the surface to release the latch; in Plug (Clear) task, E2E clones the executed gripper width and under-clamps the plug, causing it to slip during insertion, while E2C preserves the tighter clamping command. Through the position controller, these offsets act as virtual equilibrium points that generate sustained contact force.

\subsubsection{\textbf{Dual-State Conditioning for Force Perception (Q3)}}

\begin{wraptable}{r}{0.48\textwidth}
  \vspace{-14pt}
  \centering
  \caption{\textbf{Dynamic Manipulation Capability.} We report the success rates for Conveyor Weight Sorting (decomposed into Grasp, Place, Sort stages) and Dynamic Tossing tasks.}
  \label{tab:dynamic_tasks}
  \resizebox{\linewidth}{!}{
    \begin{tabular}{ll ccc c}
    \toprule
    & & \multicolumn{3}{c}{\textbf{Conveyor Weight Sorting}} & \multicolumn{1}{c}{\textbf{Tossing}} \\
    \cmidrule(lr){3-5} \cmidrule(lr){6-6} 
    \textbf{Model} & \textbf{Inference} & Grasp & Place & Sort & Success \\
    \midrule
    \multicolumn{2}{l}{\textit{Diffusion Policy (DP)}} & & & & \\
    E2E & Na\"ive Async & 36 / 50 & 33 / 36 & 17 / 33 & - \\ 
    E2C & Na\"ive Async & 44 / 50 & 36 / 44 & 20 / 36 & - \\ 
    \textbf{EC2C} & Na\"ive Async & \textbf{46 / 50} & \textbf{46 / 46} & \textbf{41 / 46} & - \\ 
    \midrule
    \multicolumn{2}{l}{\textit{$\pi_0$ (Flow Matching)}} & & & & \\
    EC2C & Sync & 0 / 50 & - & - & 14 / 20 \\
    E2E & Na\"ive Async & 35 / 50 & 32 / 35 & 16 / 32 & 17 / 20 \\ E2C & Na\"ive Async & 46 / 50 & 45 / 46 & 28 / 45 & 17 / 20 \\ \textbf{EC2C} & \textbf{Inpainting} & \textbf{50 / 50} & \textbf{49 / 50} & \textbf{48 / 49} & \textbf{18 / 20} \\ 
    \bottomrule
    \end{tabular}%
  }   
  \vspace{-10pt}
\end{wraptable}

\begin{figure}[b]
    \vspace{-10pt}
    \centering
    \begin{minipage}[t]{0.49\textwidth}
        \vspace{0pt}
        \centering
        \includegraphics[width=\linewidth]{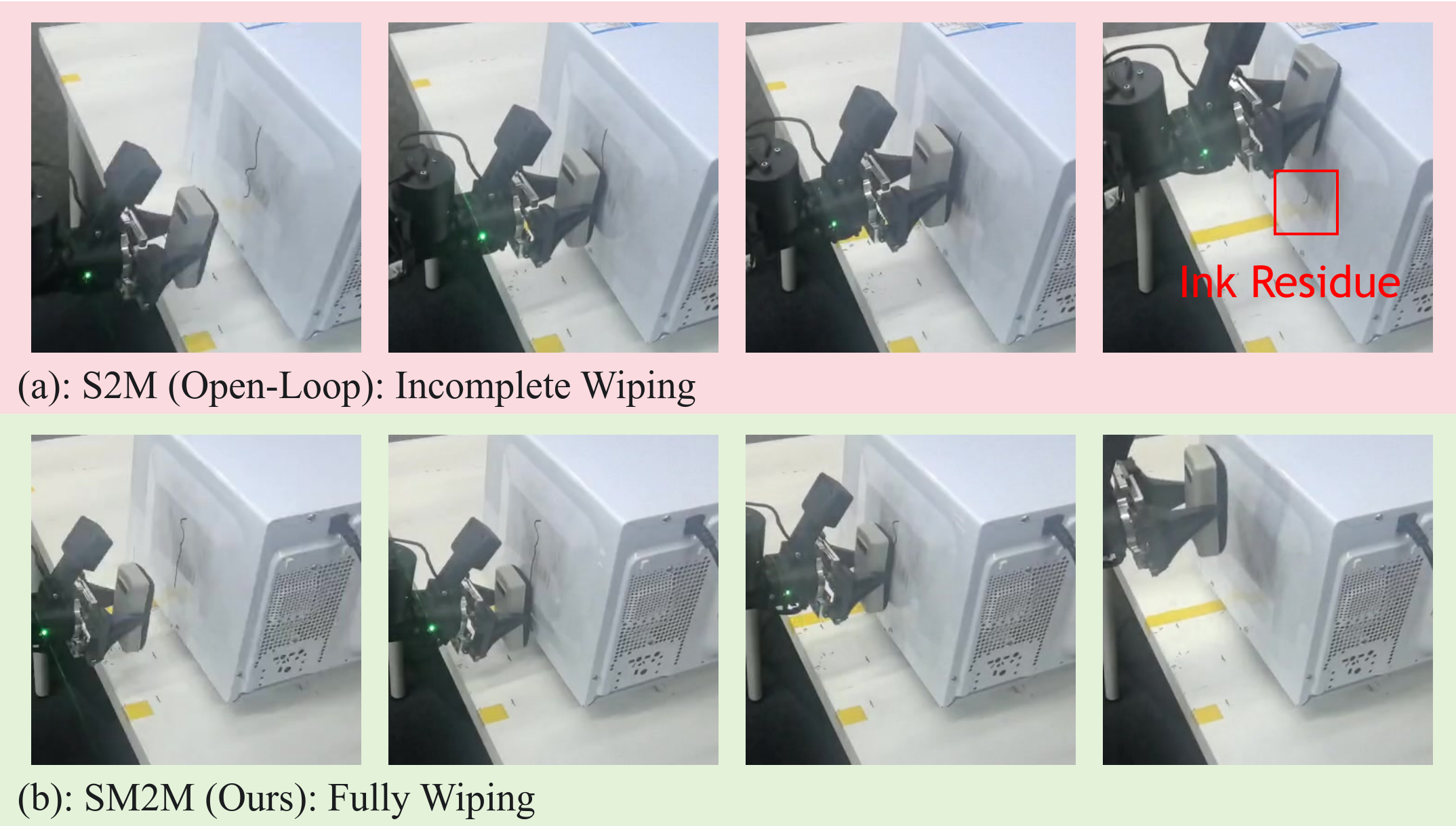}
        \caption{\textbf{Mismatch-Conditioned Wiping.} E2C leaves residue after moving with insufficient pressure; EC2C uses mismatch feedback to regulate contact and erase cleanly.}
        \label{fig:wiping_result}
    \end{minipage}
    \hfill
    \begin{minipage}[t]{0.49\textwidth}
        \vspace{0pt}
        \centering
        \includegraphics[width=\linewidth]{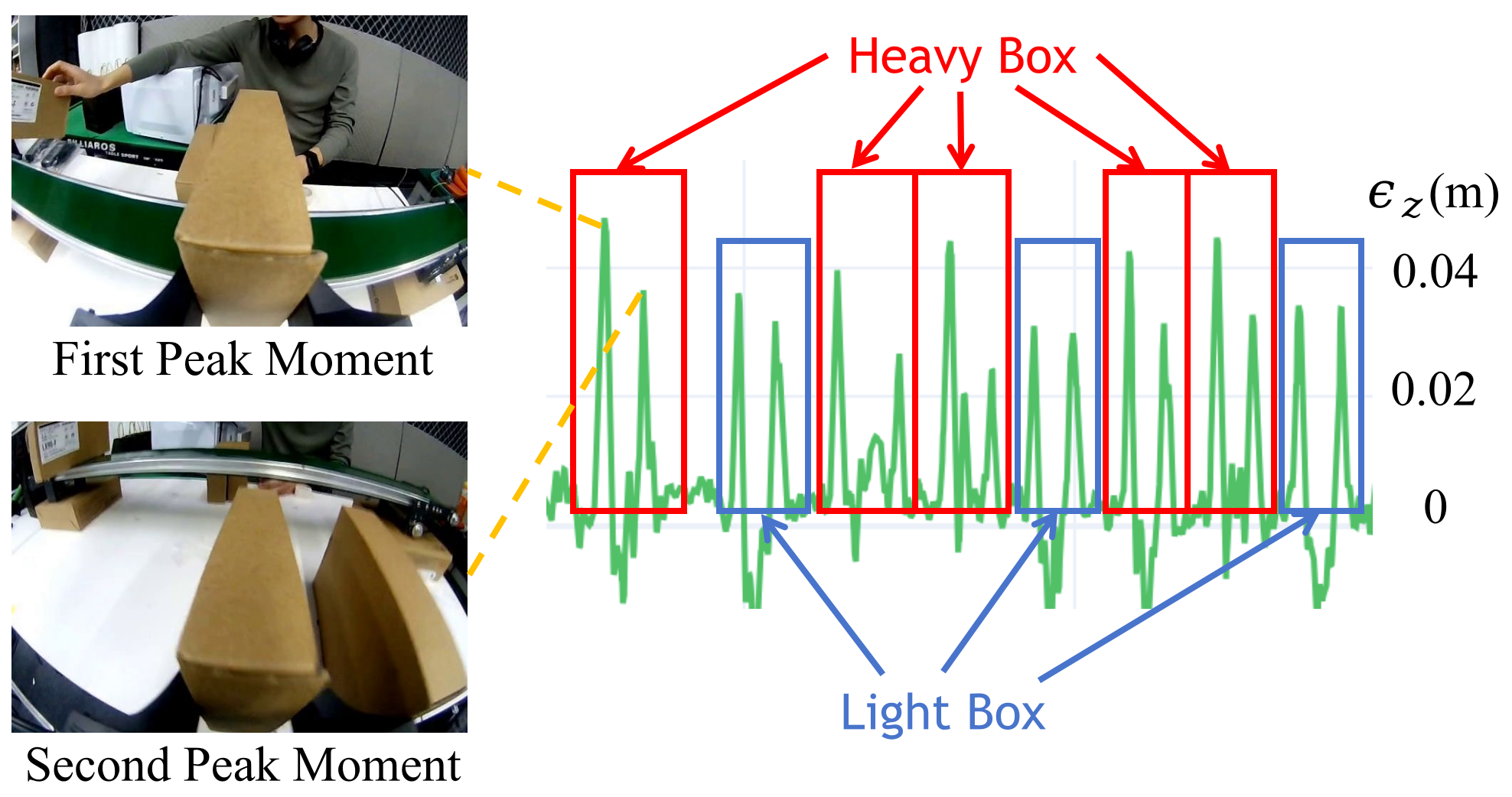}
        \caption{\textbf{Mismatch in Conveyor Sorting.} Heavy trials (red) show larger vertical command-execution mismatch peaks ($\epsilon_z$) during lifting than light trials (blue).}
        \label{fig:mismatch_visualization}
    \end{minipage}
    \vspace{0pt}
\end{figure}

As shown in Table~\ref{tab:main_results}, EC2C improves over E2C in quasi-static contact tasks where the timing of force application matters. In Blind Plug Insertion, E2C achieves $8/20$ because it often enters the high-force insertion phase while still misaligned, causing jams or plug slip; EC2C improves success to $16/20$ by distinguishing faceplate contact from sliding into the socket. In Surface Wiping, E2C often starts lateral motion before sufficient normal pressure is established, yielding $4/20$ strict success despite a $13/20$ effective rate; Fig.~\ref{fig:wiping_result} shows the resulting residue. EC2C captures the human press-then-move strategy through longitudinal mismatch and reaches $18/20$ strict success. Table~\ref{tab:dynamic_tasks} further shows that mismatch also helps in dynamic physical inference: in Conveyor Weight Sorting, where objects are visually identical, E2E and E2C sort near chance, while EC2C reaches $48/49$ by using vertical mismatch peaks during lifting as a mass cue (Fig.~\ref{fig:mismatch_visualization}).

\subsubsection{\textbf{Observation Horizon and Temporal Cues (Q4)}}

\begin{wraptable}{r}{0.3\textwidth}
  \vspace{-34pt}
  \centering
  \caption{\textbf{Observation Horizon Ablation.} Longer command-execution history helps dynamic mass inference but can hurt quasi-static contact tasks.}
  \label{tab:horizon_ablation}
  \resizebox{\linewidth}{!}{%
    \begin{tabular}{lcc}
    \toprule
    \textbf{Task} & \textbf{$T_{obs}=1$} & \textbf{$T_{obs}=9$} \\
    \midrule
    Plug (Clear) & \textbf{19/20} & 12/20 \\
    Plug (Blind) & \textbf{16/20} & 12/20 \\
    Weight Sorting & 18/30 & \textbf{48/49} \\
    \bottomrule
    \end{tabular}%
  }  
  \vspace{-10pt}
\end{wraptable}

Table~\ref{tab:horizon_ablation} shows that observation horizon is task-dependent. In quasi-static plug insertion, a longer history degrades performance. We attribute this to a history shortcut: longer windows make next-command prediction easier in training, but can bias the policy toward past trajectory rather than the current visual alignment and contact state. In contrast, Weight Sorting benefits only when the history window covers the vertical mismatch peak during lifting, which provides the mass-discriminative cue.

\subsubsection{\textbf{System-dynamics Shift (Q5)}}

\begin{wrapfigure}{r}{0.5\textwidth}
  \vspace{-40pt}
  \centering
  \includegraphics[width=\linewidth]{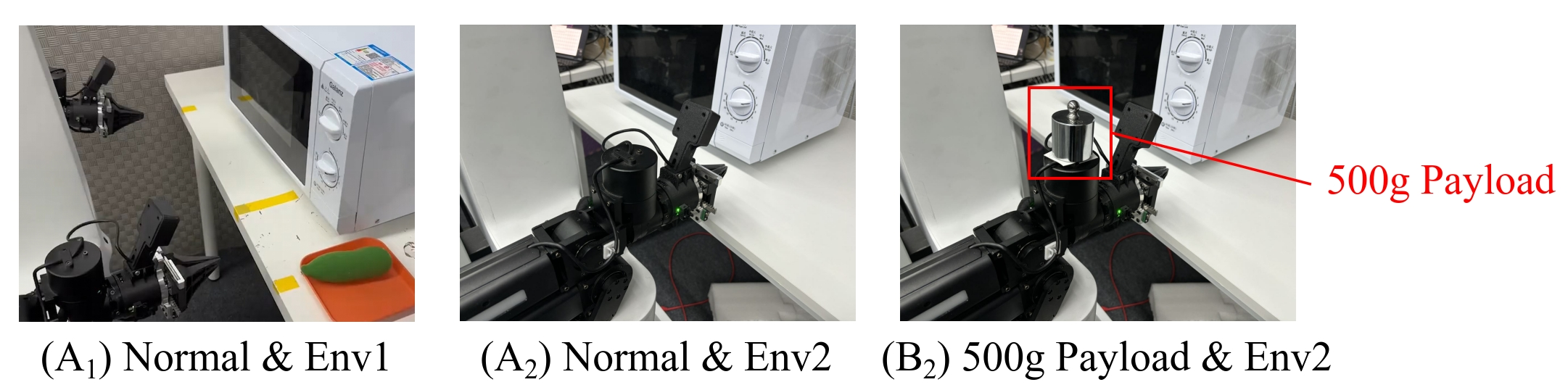}
  \caption{\textbf{System-dynamics Shift Ablation (Q5).}
  We evaluate the Microwave task under nominal system-dynamics A and 500g-payload B; subscripts 1/2 denote visual environments.}
  \label{fig:payload_plot}
  \vspace{0pt}
  
  \captionof{table}{\textbf{System-dynamics Shift Results (Q5).}}
  \label{tab:payload_generalization}
  \resizebox{\linewidth}{!}{%
    \begin{tabular}{lccccc}
    \toprule
    \textbf{Method} & \textbf{A$_1$$\rightarrow$A$_1$} & \textbf{A$_1$$\rightarrow$B$_2$} & \textbf{B$_2$$\rightarrow$B$_2$} & \textbf{A$_1$B$_2$$\rightarrow$A$_2$} & \textbf{A$_1$A$_2$B$_2$$\rightarrow$A$_2$} \\
    \midrule
    \textbf{E2C} & 18/20 & 0/10 & 7/10 & 0/10 & \textbf{10/10} \\
    \textbf{EC2C (Ours)} & \textbf{20/20} & \textbf{8/10} & \textbf{8/10} & \textbf{10/10} & \textbf{10/10} \\
    \bottomrule
    \end{tabular}%
  }
  \vspace{-12pt}
\end{wrapfigure}

Q5 (Fig.~\ref{fig:payload_plot}) tests system-dynamics generalization by attaching an unmodeled 500g payload to the robot arm during the Microwave task. As shown in Table~\ref{tab:payload_generalization}, EC2C outperforms E2C under both zero-shot payload shift (A$_1$$\rightarrow$B$_2$) and visual-dynamics recombination (A$_1$B$_2$$\rightarrow$A$_2$). This suggests that mismatch helps identify the current command-to-execution response.

\subsubsection{\textbf{Latency-Robust Asynchronous Execution (Q6)}}

\begin{wraptable}{r}{0.5\linewidth}
  \vspace{-12pt}
  \centering
  \caption{\textbf{Impact of Inference Latency.} We report the Success Rate and Motion Jerk ($m/s^3$) in Conveyor Weight Sorting Task under artificial inference delays ranging from $100$ms to $400$ms.}
  \label{tab:latency_results}
  \resizebox{\linewidth}{!}{%
    \begin{tabular}{@{}lcc cc cc cc@{}}
    \toprule
    & \multicolumn{2}{c}{\textbf{100 ms}} & \multicolumn{2}{c}{\textbf{200 ms}} & \multicolumn{2}{c}{\textbf{300 ms}} & \multicolumn{2}{c}{\textbf{400 ms}} \\
    \cmidrule(lr){2-3} \cmidrule(lr){4-5} \cmidrule(lr){6-7} \cmidrule(lr){8-9}
    \textbf{Architecture} & \textbf{Succ} & \textbf{Jerk} & \textbf{Succ} & \textbf{Jerk} & \textbf{Succ} & \textbf{Jerk} & \textbf{Succ} & \textbf{Jerk} \\ \midrule
    \textbf{Na\"ive Async} & 50/50 & 0.75 & 47/50 & 0.80 & 45/50 & 1.01 & 40/50 & 1.28 \\
    \textbf{Ours} & 50/50 & 0.75 & 48/50 & 0.78 & 48/50 & 0.82 & 45/50 & 0.88 \\ 
    \bottomrule
    \end{tabular}%
  }
  \vspace{-20pt}
\end{wraptable}

Table~\ref{tab:dynamic_tasks} shows that synchronous $\pi_0$ fails when latency changes the task state, such as moving-target grasping on Conveyor, and also hurts Tossing by breaking release continuity. Table~\ref{tab:latency_results} shows that inpainting keeps motion smoother and success higher than Na\"ive Async.

\section{Limitations and Conclusion}
\label{sec:conclusion}

\textbf{Limitations.}
Command-execution mismatch is a composite tracking-error feature affected by contact, latency, friction, gravity, payload, controller gains and kinematic configuration. Its force-correlated interpretation is most appropriate under local quasi-static; for stick-slip friction, nonlinear compliance, saturation, or configurations near singularity, it should be treated as a broader execution-grounded cue rather than calibrated force sensing.

The observation horizon is task-dependent. Long histories are useful when the task-relevant latent state cannot be identified from a single frame, as in weight sorting. For fine-grained contact-rich tasks such as wiping and plug insertion, the current observation and command-execution mismatch are often sufficient; longer histories can instead result in causal confounding. We therefore use long histories only when such temporal evidence is necessary. Finally, the learned mismatch semantics are tied to the teleoperation device, low-level controller, and embodiment. Our payload experiment provides evidence for robustness within one hardware family, but transfer across controllers, controller parameters, and embodiments remains important future work.

\textbf{Conclusion.}
This paper studies the I/O space of teleoperation-based imitation learning under imperfect low-cost controllers. E2C is already a strong and common formulation; our central finding is that paired command and execution traces should both be exposed to the policy when available. EC2C preserves the virtual-equilibrium benefit of E2C while making the command-execution mismatch available as an execution-grounded cue. On low-cost sensorless hardware, this simple design improves contact-rich manipulation across plug insertion, wiping, microwave manipulation, tossing, and conveyor weight sorting, without explicit force, tactile, or motor-current policy input.

\clearpage
\bibliography{references}

\appendix

\section{Task Details}

In this section, we provide a detailed breakdown of each manipulation task and analyze the correspondence between the physical execution stages and the recorded \emph{command-execution mismatch} ($\epsilon_t = \mathbf{x}^c_t \ominus \mathbf{x}^e_t$). The mismatch curves are plotted in the local robot base frame: X-axis (Forward), Y-axis (Left), and Z-axis (Up).

\begin{figure*}
   \centering
   \includegraphics[width=1.0\textwidth]{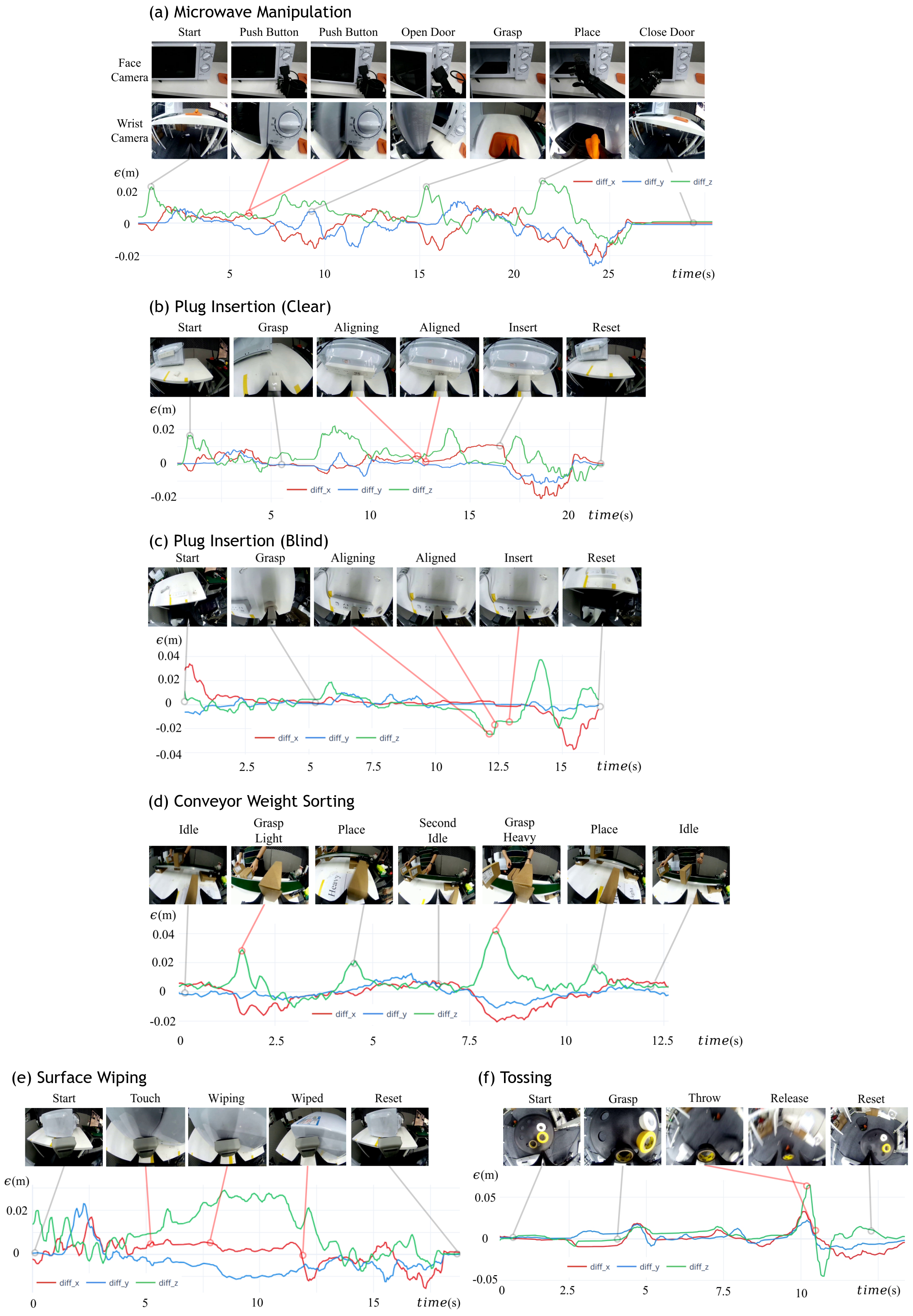}
   \caption{Task frames and command-execution mismatch visualization across tasks.}
   \label{fig:task_frames}
\end{figure*}

As illustrated in Fig.~\ref{fig:task_frames}, each task is decomposed into key semantic stages (e.g., Push Button, Aligning, Wiping). The corresponding mismatch curves show that $\epsilon_t$ is not merely random noise; it varies consistently with controller response, contact, payload, and task dynamics.

We highlight three recurring patterns in these signals across different tasks:
\begin{itemize}
   \item \textbf{Free-Space Tracking Lag:} Mismatch is not limited to contact. During rapid free-space motion (e.g., the Start phase), command and execution can differ because of controller latency, inertia, and tracking lag. These patterns show that mismatch reflects controller response even without contact, rather than serving as a contact-only signal.
   \item \textbf{Contact-Generating Offsets:} In contact-rich stages, such as Pushing the Button (Fig.~\ref{fig:task_frames} (a)) or Wiping Surface (Fig.~\ref{fig:task_frames} (e)), the mismatch often stabilizes into a sustained non-zero offset. Under local quasi-static contact, this offset is force-correlated and can indicate that the command target lies beyond the contacted surface, producing sustained contact pressure through the position controller.
   \item \textbf{Latent Physical-State Cues:} The mismatch signal also provides a proprioceptive cue for latent physical properties. In the Conveyor Weight Sorting task (Fig.~\ref{fig:task_frames} (d)), lifting a Heavy object induces a larger vertical mismatch peak than lifting a Light object due to gravity and payload inertia. The temporal mismatch profile provides evidence for inferring object mass and routing it to the correct bin.
\end{itemize}

\section{Model Architecture and Implementation Details}

In this section, we provide the implementation specifics of the Dual-State Conditioning framework and the Latency-Adaptive Inpainting mechanism. We detail how we adapt standard backbone architectures--specifically Diffusion Policy (DP) \cite{chi2023diffusion} and $\pi_0$ \cite{black2024pi0}--to incorporate the command-execution mismatch and handle inference latency.

\subsection{Input Representation: Dual-State Conditioning}

\textbf{Coordinate Reference Frames.} 
The pose of each robotic arm (both Leader and Follower) is represented in its respective Local Base Frame (the fixed coordinate system attached to the mounting base of each manipulator). For the bimanual Microwave task, the left and right arms operate in independent coordinate systems. We do not perform extrinsic calibration to unify them into a common global frame; instead, the policy implicitly learns the spatial coordination between the two independent base frames from the demonstration data. The pose $\mathbf{x} \in \mathbb{R}^7$ consists of translation $(x, y, z)$, Euler angles $(r, p, y)$, and gripper width. Note that these are \textbf{absolute poses} relative to the local base.

\textbf{Input Modalities and Temporal Structure.}
The input configuration is tailored to the task requirements. For the bimanual \textbf{Microwave} task, the policy conditions on visual observations from three RGB streams (left wrist, right wrist, and first-person view). The proprioceptive input is constructed by concatenating the states of all four arm components---Follower Left, Follower Right, Leader Left, and Leader Right---resulting in a unified \textbf{28-dimensional} vector. In contrast, for all single-arm tasks (\textbf{Plug, Wiping, Tossing, Sorting}), the input uses a single wrist-mounted camera view and a \textbf{14-dimensional} proprioceptive vector formed by concatenating the single-arm Follower and Leader states. We decouple the temporal observation horizons for these modalities: visual features are extracted from the \textbf{current frame} ($T_{img}=1$) to minimize computational load, while proprioceptive states can be provided as a \textbf{history sequence} ($T_{obs} \ge 1$) to supply task-dependent temporal context for interpreting command-execution mismatch.

\textbf{Robustness to Task-Irrelevant Inputs.}
Initial pilot experiments suggested that including task-irrelevant modalities, such as adding the left-arm state or secondary camera views to single-arm manipulation tasks, can degrade success rates. We attribute this to the model attempting to identify spurious correlations within these high-dimensional nuisance variables. Consequently, we explicitly tailor the input observation space to the degrees of freedom and visual field required for each task. This keeps the policy focused on the relevant inputs and improves robustness on low-cost hardware.

\subsection{Backbone Architecture Modifications}

\textbf{Adaptation for $\pi_0$ (Flow Matching).}
For the $\pi_0$ architecture, which utilizes a Transformer backbone, we treat each timestep of the trajectory as a single input token. To align with the pre-trained embedding layer, which expects a fixed dimension of $D=32$, we zero-pad the unified 14-dimensional or 28-dimensional proprioceptive vectors defined above to the required size. A critical implementation detail lies in constructing tokens for the \textit{committed command} (the future leader trajectory fixed by the inpainting mask). Since the future follower execution is unknown at inference time, we construct these future tokens by concatenating the future leader state with the current follower state during both training and inference. This avoids using unavailable future execution states and keeps the training input consistent with inference.

\textbf{Adaptation for Diffusion Policy (DP).} We utilize the standard CNN-based 1D Temporal U-Net as our backbone. To implement \textit{Latency-Adaptive Inpainting}, we modify the input to the noise prediction network by channel-wise concatenating the noisy action $\mathbf{A}_{t+1:t+H}$ (where $H$ denotes the prediction horizon), the committed command $\mathbf{X}^c_{cond}$, and the binary validity mask $\mathbf{M} \in \{0, 1\}^H$. Crucially, $\mathbf{X}^c_{cond}$ is constructed by taking the committed trajectory for the first $k$ steps and \textit{padding the remaining $H-k$ steps with the last committed state} $\mathbf{x}^c_{t+k}$. This ensures the network receives a continuous guidance signal rather than zero-filled voids. Furthermore, we employ \textit{asymmetric observation horizons}: the visual encoder processes only the current frame ($T_{img}=1$) to minimize latency, while the proprioceptive encoder receives a history window ($T_{obs}$) of dual-state vectors $[\mathbf{x}^c, \mathbf{x}^e]$ when temporal context is required.

\subsection{Training Objective and Inference Execution}

\textbf{Minimalist Backbone Integration.} 
We emphasize that our framework relies solely on input conditioning, requiring \textbf{no modifications} to the core training objectives, loss functions, or the internal sampling dynamics (e.g., noise scheduling or inference sampler) of the backbone models. For both Diffusion Policy (utilizing standard MSE loss with DDIM scheduling) and $\pi_0$ (utilizing the standard Flow Matching objective with ODE solvers), the optimization targets remain invariant. This alignment ensures that our Dual-State Conditioning framework serves as a plug-and-play module, easily integrated into existing robot learning pipelines by simply modifying the input encoder without the need to alter the underlying generative mechanics.

\textbf{Latency-Aware Execution Strategy.} 
During inference, the policy predicts a full action chunk of length $T_{pred}$. However, due to the non-zero inference latency $\delta_{inf}$, the first $k = \lceil \delta_{inf} / \Delta t \rceil$ steps of the prediction correspond to the time interval that has already elapsed during computation. We \textbf{discard} these first $k$ steps and execute the trajectory starting from step $k+1$. In the context of our inpainting framework, these discarded steps correspond precisely to the committed command used as input conditioning.

\section{Training Details and Hyperparameter Analysis}

In this section, we provide the training protocols and a detailed analysis of the hyperparameter choices. We specifically discuss the trade-off between history length and causal confusion in contact-rich manipulation.

\subsection{Experimental Setup and Hyperparameters}

All models were trained on a single \textbf{NVIDIA A100 (80GB) GPU}. The training configuration is standardized across tasks to ensure fair comparison, with specific variations only in the observation history and inference strategy to accommodate task dynamics.

The detailed hyperparameters, primarily based on the $\pi_0$ configuration, are listed in Table ~\ref{tab:hyperparams}.

\begin{table}[h]
\centering
\caption{Standardized Training Hyperparameters}
\label{tab:hyperparams}
\begin{tabular}{ll}
\toprule
\textbf{Category} & \textbf{Value / Setting} \\
\midrule
\multicolumn{2}{l}{\textit{Architecture ($\pi_0$ Backbone)}} \\
Base Model & Paligemma-3B (Gemma-2B + SigLIP) \\
Action Expert & 300M Parameter MLP \\
Precision & \texttt{bfloat16} \\
\midrule
\multicolumn{2}{l}{\textit{Optimization}} \\
Optimizer & AdamW ($\beta_1=0.9, \beta_2=0.95$) \\
Peak Learning Rate & $2.5e^{-5}$ \\
LR Schedule & Warmup ($1,000$ steps) + Cosine Decay \\
Weight Decay & $1e^{-10}$ \\
Batch Size & 16 \\
Training Steps & 30,000 \\
\midrule
\multicolumn{2}{l}{\textit{Common Prediction Parameters}} \\
Prediction Horizon ($T_{pred}$) & 30 steps \\
Execution Horizon ($T_{exec}$) & 15 steps (Open-loop duration) \\
Visual Input ($T_{img}$) & 1 (Current Frame Only) \\
\midrule
\multicolumn{2}{l}{\textit{Task-Specific Variations}} \\
\textbf{Quasi-Static Tasks} & $T_{obs}=1$, Inference: \textit{Synchronous} \\
{\small (Microwave, Plug, Wipe)} & \\
\textbf{Dynamic Tasks} & $T_{obs}=9$, Inference: \textit{Inpainting} \\
{\small (Sorting, Tossing)} & \\
\bottomrule
\end{tabular}
\end{table}

\subsection{Hyperparameter Analysis}

\textbf{The Risk of Causal Confusion ($T_{obs}$).}
A key finding in our experiments is that ``more history is not always better.''
\begin{itemize}
    \item \textbf{Quasi-Static Tasks ($T_{obs}=1$):} For tasks like \textit{Surface Wiping} or \textit{Plug Insertion}, the current observation and command-execution mismatch are often sufficient for interpreting the immediate geometric and contact state. We found that conditioning on a long history ($T_{obs}=9$) degraded success rates. We attribute this to \textbf{causal confounding}: longer windows can bias the policy toward past trajectory trends, task phase, or operator compensation rather than the current alignment and contact condition. Therefore, we use $T_{obs}=1$ for these tasks.
    \item \textbf{Dynamic Tasks ($T_{obs}=9$):} In contrast, for \textit{Weight Sorting}, the instantaneous state is insufficient to distinguish object mass. The policy requires a temporal window that covers the vertical mismatch peak during lifting, which provides the mass-discriminative cue. Thus, we use $T_{obs}=9$ for dynamic tasks where temporal accumulation is physically meaningful.
\end{itemize}

\textbf{Action Execution Horizon ($T_{exec}$).}
The execution horizon determines the open-loop control duration before replanning. Our pilot studies suggested that high-precision tasks (e.g., \textit{Blind Plug Insertion}) benefit from a shorter horizon ($T_{exec} \approx 10$) to increase replanning frequency and contact reactivity. In contrast, friction-heavy tasks (e.g., \textit{Wiping}) benefit from a longer horizon ($T_{exec} \approx 20$) to reduce stagnation by committing to smoother, longer motion chunks. Despite these task-specific optima, we selected a unified \textbf{$T_{exec}=15$} in all reported benchmarks for a fair comparison. We also perform $3\times$ temporal interpolation on the predicted action chunk. The standardized execution horizon of $T_{exec}=15$ model steps corresponds to a physical duration of 0.75 seconds at a 60Hz control frequency.

\section{Failure Analysis}

\begin{figure*}[thbp]
   \centering
   \includegraphics[width=\textwidth]{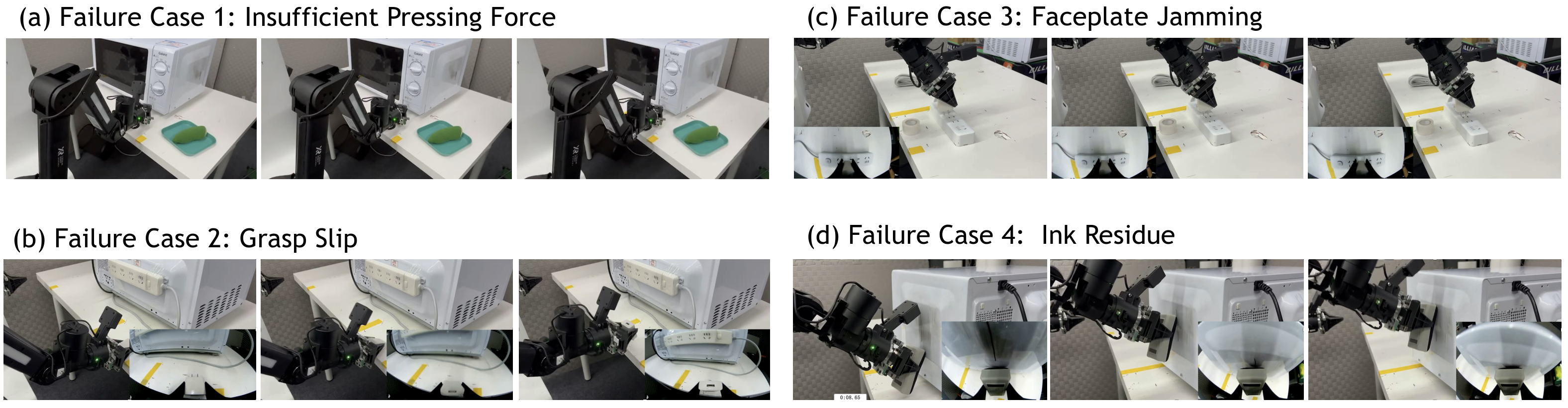}
   \caption{Failure cases in different tasks.}
   \label{fig:failure_cases}
\end{figure*}

We analyze failure modes to highlight the roles of preserving command offsets (E2C) and conditioning on execution response (EC2C). Fig.~\ref{fig:failure_cases} contrasts the contact-pressure deficit of E2E with the execution-response deficit of the open-loop E2C baseline.

\textbf{(a) Button Press Failure (E2E, Microwave).}
\textit{Failure of Force Generation.} The E2E policy accurately tracks the geometric trajectory but halts exactly at the button surface. Lacking the virtual penetration intent, it fails to generate the activation force ($\approx 10\text{N}$), causing the robot to freeze at the button.

\textbf{(b) Grasp Slip (E2E, Plug Clear).}
\textit{Failure of Grip Force.} By cloning the \textit{measured} gripper width, the E2E policy outputs a zero-margin command. This results in negligible clamping force, causing the rigid plug to slip or rotate during transport.

\textbf{(c) Faceplate Jamming (E2C, Plug Blind).}
\textit{Missing Execution Response.} While E2C generates sufficient insertion force, it operates open-loop. It cannot distinguish between ``sliding into the hole'' and ``hitting the faceplate''. Consequently, it often commits to a high-force insertion while misaligned, leading to jamming against the rigid surface rather than correcting its pose.

\textbf{(d) Ink Residue (E2C, Wiping).}
\textit{Failure of Contact Coordination.} Effective surface wiping requires a specific coordination strategy: establish normal force first, then initiate lateral motion. E2C often misses spots or moves too quickly to ensure complete erasure.

\end{document}